\documentclass{article}
\usepackage[preprint]{iclr2027_conference}
\usepackage{times}

\usepackage{amsmath}
\usepackage{amssymb}
\usepackage{mathtools}
\usepackage{graphicx}
\usepackage{tikz}
\usepackage{booktabs}
\usepackage{multirow}
\usepackage{placeins}
\usepackage{xcolor}
\usepackage{microtype}
\usepackage[hidelinks]{hyperref}
\usepackage{url}

\title{\methodname{}: Learning Robot Policies on Physical Time}

\author{Changhao Wang \quad Jaden Clark \quad Shuran Song\\
Stanford University}

\newcommand{\methodname}{Chronos Policy}
\newcommand{\methodshort}{Chronos Policy}

\begin{document}
\maketitle

\begin{abstract}
Robot policies often convert sensing and control into fixed-rate observation--action sequences, leaving physical time implicit in sequence position. This assumption breaks when observation rates or delays vary, or when inference latency changes. We introduce \methodname{}\footnote{``Chronos'' is the Greek word for time, reflecting that the policy reasons directly over physical time.}, which makes physical time explicit by conditioning on observation capture times and desired action execution times. We study two ways to incorporate time into a diffusion transformer: adding timestamp embeddings to token features, or encoding physical time in positional encoding through time-aware rotary positional encodings (time-aware RoPE). We find that time-aware RoPE is more robust, allowing attention to depend directly on physical time while keeping token content unchanged. During training, we temporally randomize which observations are sampled and when the action chunk begins, exposing the policy to a wider range of sensing and execution schedules that may occur during deployments. This also enables learning from asynchronous, multi-rate data without forcing all measurements onto a fixed temporal grid. Across simulation and real-world experiments, \methodname{} improves performance on dynamic tasks, where observation-to-action timing is critical, and on multisensory tasks, where the relative timing between observations matters. On standard quasi-static tasks, it preserves nominal performance while improving robustness to changes in sensing and inference timing. Project website: \url{https://changhaowang.github.io/chronospolicy/}.
\end{abstract}


\section{Introduction}
\label{sec}

Real robot systems rarely run on a fixed clock. Consider a robot grasping a moving object from a conveyor belt: the same displacement between two images can correspond to very different object velocities depending on whether the images were captured 20 ms or 200 ms apart. Likewise, the appropriate grasp depends on when the predicted action will actually be executed. Successful control therefore depends not only on \emph{what} the robot observes, but also on \emph{when observations are captured} and \emph{when actions execute}.

\begin{figure}[t]
\centering
\includegraphics[width=\textwidth]{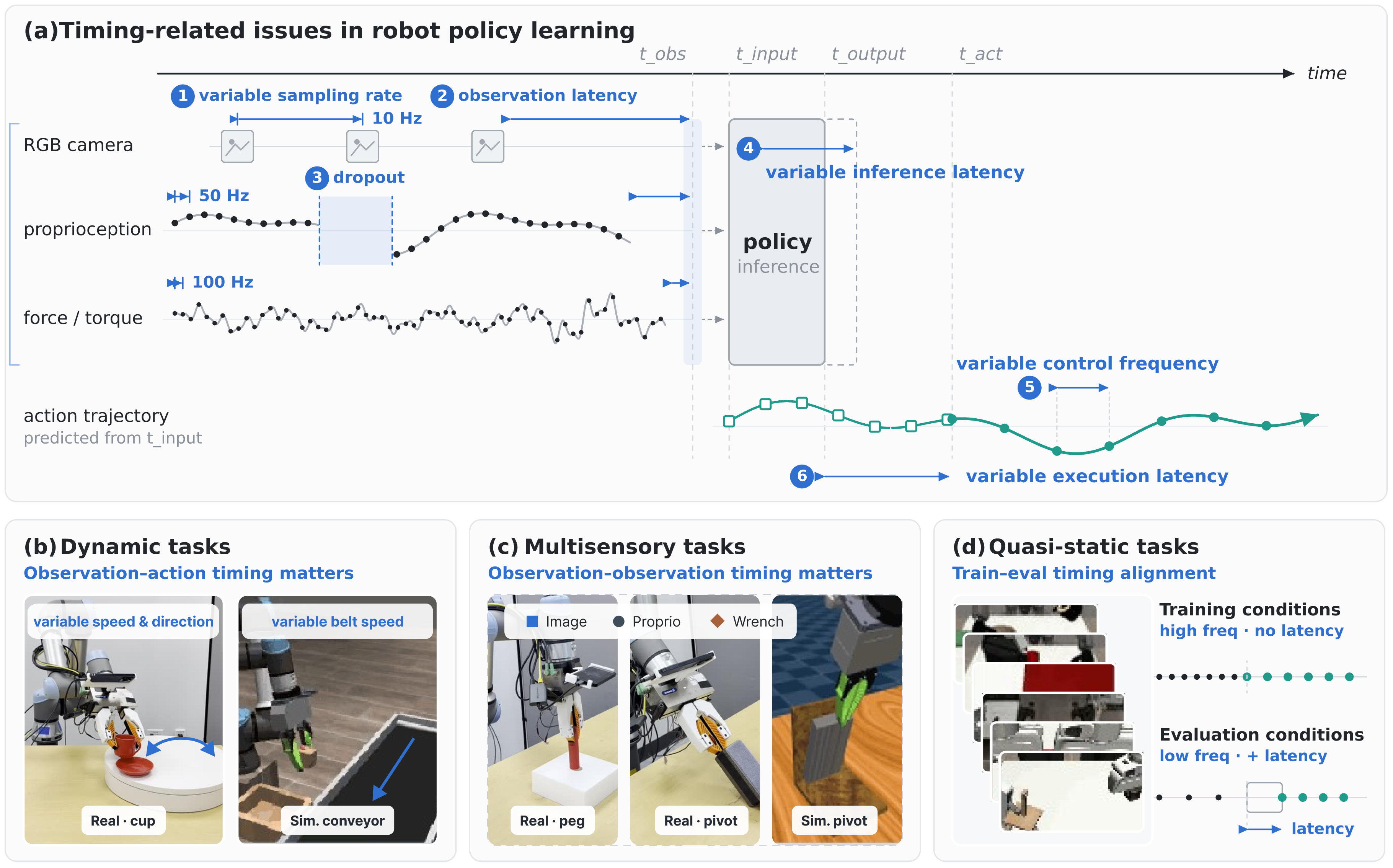}
\caption{\textbf{Timing-related issues in robot policy learning.}
(a) Real robot systems exhibit multiple sources of temporal variation, including variable sensor sampling rate, observation latency and dropout, inference latency, control frequency, and execution latency. These timing effects matter differently across tasks. (b) In dynamic tasks, the key temporal relationship is between \emph{when an observation is captured} and \emph{when an action takes effect}. (c) In multisensory tasks, the key temporal relationship is between \emph{when different observations are captured}. (d) Even in quasi-static tasks, differences between data-collection and deployment timing can create temporal mismatch. Standard fixed-rate policy representations obscure these distinctions by treating sequence position as a proxy for time.}
\label{fig:teaser}
\end{figure}

More generally, robot learning systems are exposed to timing variation at every stage of sensing and control. Sensors operate at different rates and latencies, inference takes time, controllers may execute at different frequencies. Yet many imitation learning methods convert demonstrations into fixed-rate observation--action sequences, using sequence position as a proxy for physical time. This assumes a consistent relationship between sequence position and elapsed time; when that relationship changes, the policy cannot tell how much time has passed between observations or how far in the future an action will execute. Prior work has addressed specific deployment-time effects, such as latency matching~\citep{chi2024universal} and asynchronous action-chunk execution~\citep{black2026real}. We take a different approach: instead of treating timing variation as an external mismatch to correct, we represent it explicitly as part of the policy input.

Concretely, we formulate the policy as:
\begin{equation}
\mathbf A_{t_a} = f_\theta(\mathbf O_{t_o}, \mathbf t_o, \mathbf t_a).
\label{eq}
\end{equation}
Here, $t_o$ specifies when observations were captured and $t_a$ when actions should be executed. Their relative timing represents observation spacing, measurement age, and the requested execution schedule.

A key design choice is how this timing information should enter the policy. One option is to encode timestamps and add them directly to the token embeddings, mixing \emph{what} a token represents with \emph{when} it occurs. We instead find it more effective to keep the token content unchanged and use physical time only as positional encoding in attention. \methodname{} therefore replaces index-based positional encoding with \emph{time-aware rotary positional encodings} (time-aware RoPE): observation tokens use their capture times $t_o$, while noisy action tokens use their queried execution times $t_a$ as rotary coordinates. Attention then depends on elapsed time $t_i-t_j$, directly representing observation-to-observation, observation-to-action, and action-to-action timing while remaining invariant to the choice of clock origin.

Timestamps also allow us to vary the timing of training examples. We randomly change which observations are selected and when the action chunk starts. Because the policy sees these timestamps, it can correctly interpret each sampled sequence even when its timing differs from the nominal training setup. This exposes the policy to timing variations that may occur at deployment and lets it use measurements that fixed-rate downsampling would otherwise discard.

We evaluate \methodname{} across three types of tasks: dynamic tasks, where observation-to-action timing affects the correct control; multisensory tasks, where the relative timing between observations matters; and standard quasi-static tasks, where precise timing is less critical. Across simulation and real-robot experiments, \methodname{} improves robustness to sensing and inference variation while preserving performance on quasi-static tasks where timing matters less.

Our contributions are threefold:
\begin{itemize}[leftmargin=3mm]
\item We formulate robot policy learning in physical time by conditioning on observation capture times and queried action execution times.
\item We introduce \methodname{}, a diffusion transformer with time-aware RoPE that learns directly from asynchronous, multi-rate observations, without aligning them to a fixed temporal grid.
\item We develop benchmarks that implement realistic timing variation and show that \methodname{} improves dynamic manipulation through better observation--action timing and multisensory manipulation through better observation--observation timing, while preserving nominal performance on standard tasks and improving robustness to deployment-time mismatch.
\end{itemize}

\section{Related Work}
\label{sec:related}
\paragraph{Asynchronous multisensory learning.}

Multisensory observations provide feedback that enables imitation learning policies to succeed on contact-rich  tasks where vision alone is insufficient \citep{li2022see, wang2022safe, zhang2022learning, zhang2023efficient, hou2025adaptive,yin2026osmo, choi2026wild, clark2026multisensory}.
However, heterogeneous sensors often operate at different sampling rates, motivating methods that explicitly account for these differing temporal resolutions. Reactive Diffusion Policy combines slow visual planning with fast tactile feedback using separate modules~\citep{xue2025reactive}, while ImplicitRDP integrates both streams within a single network~\citep{chen2026implicitrdp}. ManipForce uses frequency-aware embeddings and cross-attention to fuse asynchronous RGB and force/torque streams~\citep{lee2025manipforce}. LAG-Fusion composes modality-specific diffusion policies at their native inference rates, rebasing delayed guidance before fusion~\citep{he2026asynchronous}, and FA-RDP adapts inference frequency to the manipulation phase~\citep{zhuo2026fardp}. These approaches establish the importance of preserving high frequency feedback.

\paragraph{Inference delay and chunk continuity.} 

Inference delay creates a mismatch between when observations are captured, when a policy predicts an action, and when that action is executed~\citep{xiao2020thinking,chi2024universal}. For action-chunk policies, this also raises a consistency problem between newly predicted actions and commands already committed for execution. Prior work addresses this through consistency-based decoding~\citep{liu2024bidirectional}, overlapping inference and execution with prefix inpainting~\citep{black2026real}, training-time conditioning on committed actions~\citep{black2025training}, masked action generation~\citep{wang2026discretertc,wang2026remac}, or predicting the future execution-time state~\citep{tang2025vlash}. Most closely related, Delay-Aware Diffusion Policy conditions on measured inference delay and constructs delay-compensated targets~\citep{liao2025delay}. Another line of work reduces inference latency by changing the diffusion process itself. Diffusion Forcing introduces independent noise levels for individual sequence positions, enabling flexible partial denoising schedules~\citep{chen2024diffusionforcing}, while Streaming Diffusion Policy applies variable noise levels across the action horizon to generate near-term actions with fewer denoising steps and lower latency~\citep{hoeg2025streaming}. These approaches reduce the timing mismatch by making policy inference faster, while \methodshort{} makes the remaining sensing and execution timing explicit to the policy; the two directions are complementary.

\paragraph{Explicit time in robot control.}
In addition to observation and inference delay, prior work has also made time
an explicit variable in robot control. Locomotion policies have conditioned on
the time available to reach a target, allowing behavior to adapt to temporal
requirements~\citep{rudin2022advanced}. More recently, manipulation policies
have been conditioned on remaining task time and a time-ratio signal, enabling
a single policy to vary its execution strategy across different temporal
regimes~\citep{jia2025time}. Other work makes control timing itself adaptive:
CTCO selects continuous options with variable durations
~\citep{karimi2023dynamic}, while TARC jointly predicts actions and their
application durations to vary control frequency online
~\citep{sukhija2025tarc}. These works use time to control the speed or duration of behavior. Chronos Policy instead represents when observations are captured and when actions are executed. This gives a more general and natural formulation that covers observation rate, sensor delay, cross-modal timing, and action execution delay within a single policy.

\paragraph{Continuous-time representations and attention.}
Explicit time representations are well established for irregular time series. mTAN learns continuous-time embeddings for attention over irregular observations~\citep{shukla2021multi}, and ContiFormer combines continuous dynamics with transformer attention~\citep{chen2023contiformer}. RoPE introduces relative positional dependence through rotations of attention queries and keys~\citep{su2024roformer}. Recent video models have similarly moved beyond fixed sequence indices: variable-frame-rate video models adapt positional representations to changing temporal spacing~\citep{zhong2026vfrtok}, while VideoRoPE introduces adjustable temporal spacing within rotary embeddings for video~\citep{wei2025videorope}. Most closely related, BulletTime conditions video generation on continuous world-time coordinates through time-aware rotary embeddings~\citep{wang2026bullettime}. \methodshort{} brings this continuous-time view to robot imitation learning, jointly representing the timestamps of heterogeneous observations and the requested execution times of actions. 

\section{Method}
\label{sec:method}

\begin{figure*}[t]
    \centering
    \includegraphics[width=\textwidth]{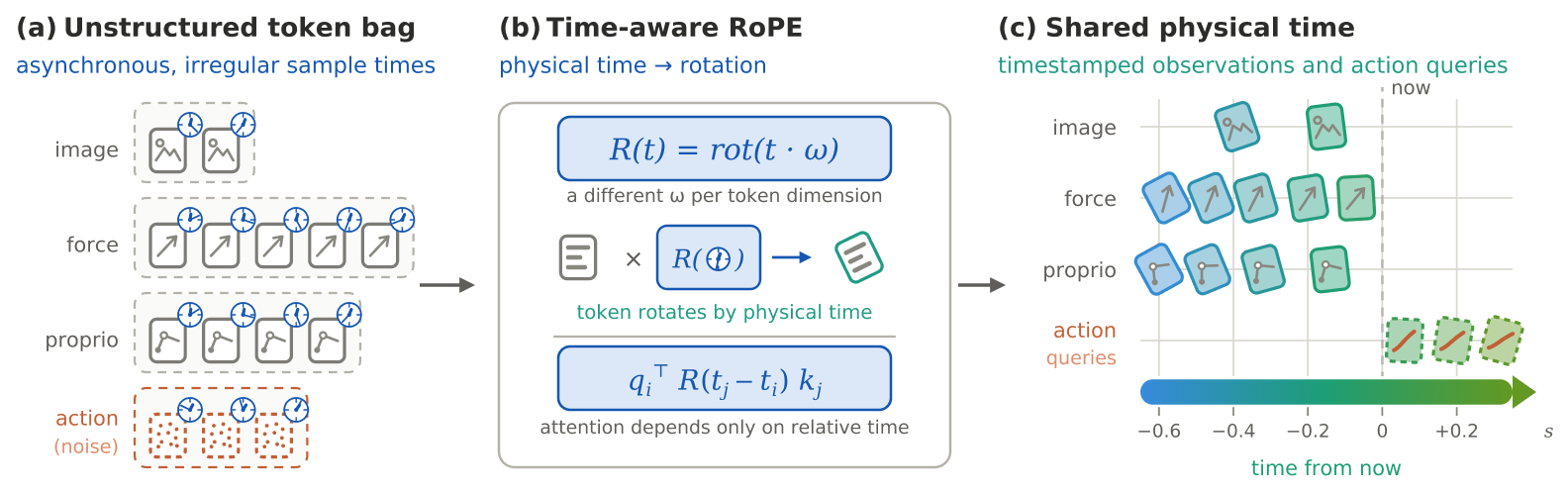}
    \caption{\textbf{Overview of \methodname{}.}
    Tokens are paired with their physical times. Time-aware RoPE uses these physical timestamps as positional coordinates, so attention depends on elapsed time rather than sequence index. The diffusion transformer then predicts an action chunk at the requested execution times.}
    \label{fig:method}
\end{figure*}

\methodname{} makes observation and action timing explicit in the policy. Each observation is associated with when it was captured, and each predicted action with when it should execute. We implement this interface with a diffusion transformer using \emph{time-aware RoPE}, which introduces time through attention while leaving the token embeddings unchanged. During training, we also vary observation and action timing while preserving the corresponding timestamps, so the policy learns to interpret the timing changes it may encounter at deployment.

\subsection{Policy Formulation on Physical Time}
\label{sec:interface}

Let $\mathbf O_{\mathbf t_o}$ denote observations captured at times $\mathbf t_o$. At time $t$, we query an $H$-step action chunk at
\begin{equation}
t_{a,j}=t+d_t+j\Delta t_a,
\qquad j=0,\ldots,H-1,
\label{eq:queries}
\end{equation}
where $d_t$ is the delay before the first queried action and $\Delta t_a$ is the action interval. The policy is therefore
\begin{equation}
\mathbf A_{\mathbf t_a}
\sim
\pi_\theta\!\left(
\cdot \mid
\mathbf O_{\mathbf t_o},
\mathbf t_o,
\mathbf t_a
\right).
\label{eq:interface}
\end{equation}

This formulation makes the timing information explicit. The spacing of $\mathbf t_o$ captures observation rate, the gap between observation and action times captures measurement delay and execution latency, and $\mathbf t_a$ specifies the desired execution frequency. For multimodal sensing, each modality simply keeps its own capture times rather than being aligned to a common sampling grid.

\subsection{Encoding Physical Time in the Policy}
\label{sec:time_attention}

Given explicit timestamps in the policy inputs, we must decide how temporal information should enter the network. One option is to encode each timestamp and add it directly to the corresponding token embedding, allowing time to modify the token representation itself. An alternative is to keep the token embedding unchanged and use time only through the positional encoding in attention. We compare these two choices in Section~\ref{sec:ablate}; the latter is consistently more robust to temporal variation. We therefore adopt \emph{time-aware RoPE}.

Every token is assigned a physical timestamp relative to the policy request time $t$. Image patches inherit the capture time of their frame, proprioceptive and force/torque tokens use their measurement times, and action tokens use their queried execution times. Thus, observation timestamps describe when measurements were captured relative to the request, while action timestamps lie in the future according to the requested execution schedule. Modality and spatial embeddings continue to represent token content, while physical time enters separately through the rotary positional encoding.

We express all timestamps relative to the policy request time,
\begin{equation}
\tilde t_i = t_i - t.
\end{equation}
For action tokens, this gives
$\tilde t_{a,j}=d_t+j\Delta t_a$, so inference latency directly shifts the queried action times into the future. We use $\tilde t_i$ as the rotary coordinate:
\begin{equation}
\widetilde q_i = R(\tilde t_i)q_i,
\qquad
\widetilde k_i = R(\tilde t_i)k_i.
\label{eq:rope}
\end{equation}
The resulting query--key interaction depends only on elapsed physical time,
\begin{equation}
\widetilde q_i^\top \widetilde k_j
=
q_i^\top R(t_j-t_i)k_j.
\label{eq:relative}
\end{equation}

Thus, attention depends directly on elapsed physical time $t_j-t_i$. Two observations separated by 20\,ms and 200\,ms receive different temporal relationships even if they occupy the same sequence positions. At the same time, shifting all timestamps by the same amount leaves the representation unchanged. Compared with adding time directly to token embeddings, time-aware RoPE keeps token content and timing separate while making pairwise temporal relationships explicit in attention.

\subsection{Training with Temporal Randomization}
\label{sec:training}

During training, we expose the policy to different sensing and execution timings through \emph{temporal randomization}. For each demonstration, we randomly vary the observation rate and delivery delay of each modality. We then select the measurements that would be available at that request time, while keeping their original capture timestamps. We also randomize when the predicted action chunk should begin. Sampling the start delay $d_t$ gives the queried execution times in Eq.~\ref{eq:queries}.

Together, these two randomizations expose the policy to temporal settings that may occur at deployment. They also allow different measurements from the recorded trajectory to be used across training examples, including measurements that a fixed-rate downsampling scheme would otherwise discard.

The timestamps are what make this possible. After randomization, the same token positions can correspond to different physical intervals. Without explicit time, these examples would be ambiguous. In \methodshort{}, the observation values $\mathbf O_{\mathbf t_o}$ enter the network as token content, while $\mathbf t_o$ and $\mathbf t_a$ are used only as temporal coordinates in time-aware RoPE. Thus, the timestamps tell the transformer \emph{when} each observation was captured and \emph{when} each action is requested without being concatenated to the token features. Changing $\mathbf t_a$ therefore changes the execution times being queried, allowing the same policy to predict actions for different future times.

We train \methodshort{} with the standard diffusion objective. Given a target action chunk $\mathbf A^*_{\mathbf t_a}$, we sample diffusion step $k$ and noise $\boldsymbol{\epsilon}$,
\begin{equation}
\mathbf A^{(k)}
=
\alpha_k \mathbf A^*_{\mathbf t_a}
+
\sigma_k \boldsymbol{\epsilon},
\qquad
\boldsymbol{\epsilon}\sim\mathcal N(0,I),
\end{equation}
and train the denoiser conditioned on the observations and queried action times:
\begin{equation}
\mathcal L(\theta)
=
\mathbb E\!\left[
\ell\!\left(
D_\theta(
\mathbf A^{(k)}, k,
\mathbf O_{\mathbf t_o},
\mathbf t_o,
\mathbf t_a),
\mathbf y^{(k)}
\right)
\right].
\label{eq:diffusion_loss}
\end{equation}
Here, $\mathbf y^{(k)}$ denotes the diffusion prediction target. We use clean-action prediction with an $\ell_1$ loss in simulation and real world.

\subsection{Using Physical Time at Deployment}
\label{sec:execution}

At deployment, we use the same physical-time interface as during training. Each available observation retains its capture time $\mathbf t_o$, while the noisy action tokens are assigned queried execution times $\mathbf t_a$. These action times enter time-aware RoPE during denoising, so when action tokens attend to observations or to one another, the attention depends on the corresponding observation-to-action and action-to-action time differences. Changing $\mathbf t_a$ therefore changes the actions predicted by the policy.

During deployment, we execute action chunks asynchronously: while the next chunk is being predicted, the robot continues executing the previous one. In simulation, the action start delay $d_t$ is set from the injected inference latency. On the real robot, the current inference time is not known in advance, so we estimate $d_t$ from previously completed requests. When a new prediction arrives, actions whose queried execution times have already passed are discarded, and the remaining actions are executed according to $\mathbf t_a$.

\section{Experiments}
\label{sec:result}

We structure our experiments around three questions. (1) \emph{when is time information useful?} We study dynamic tasks, where the correct action depends on how the environment changes over time, and multisensory tasks, where the timing between different observations matters. (2) \emph{which parts of the time-aware policy matter?} We ablate the time representation and temporal randomization used during training. (3) \emph{does adding time hurt tasks where timing is less important?} We evaluate on standard quasi-static manipulation tasks to test whether \methodname{} preserves nominal performance while improving robustness to timing changes at deployment.

\subsection{Tasks and Evaluation Setup}
\label{sec:simulation}

We evaluate \methodname{} in three types of tasks. \textbf{Conveyor} is a dynamic pick-and-place task with randomized belt speed, where the policy must account for object motion and the time between sensing and action execution. \textbf{Pivoting} is a contact-rich multisensory task that combines vision, proprioception, and force/torque measurements acquired at different rates and times. We also evaluate on six standard \textbf{quasi-static manipulation} tasks to test whether explicit time modeling preserves performance when timing is less critical.

We evaluate each policy under different timing conditions. \textbf{Nominal} uses the timing seen during training. We then vary \textbf{observation timing} by changing sensing frequency and delay, and vary \textbf{inference latency} by changing the delay before predicted actions are executed. Finally, the \textbf{combined} setting varies both observation timing and inference latency, providing a stronger test of robustness to temporal mismatch.

\begin{figure*}[t]
\centering
\includegraphics[width=\textwidth]{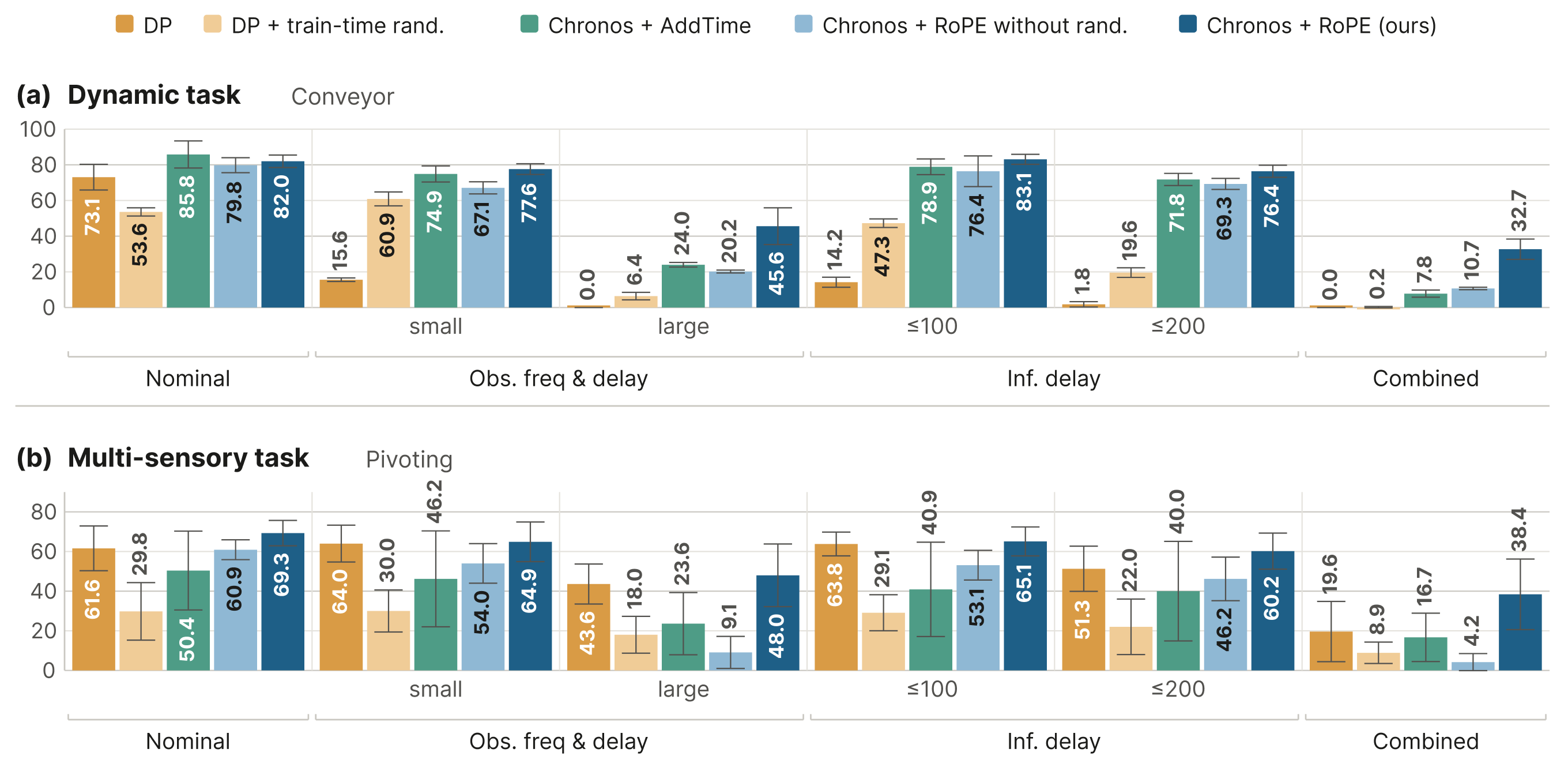}
\caption{\textbf{Simulation ablation of Chronos Policy design.}
We compare temporal randomization, additive time encoding, and time-aware RoPE to identify an effective recipe for time awareness. Temporal randomization is most effective when paired with explicit timestamps, while time-aware RoPE provides the strongest robustness across Conveyor and Pivoting.}
\label{fig:ablation}
\end{figure*}

\subsection{When Is Time Information Useful?}

\begin{figure*}[t]
\centering
\includegraphics[width=\textwidth]{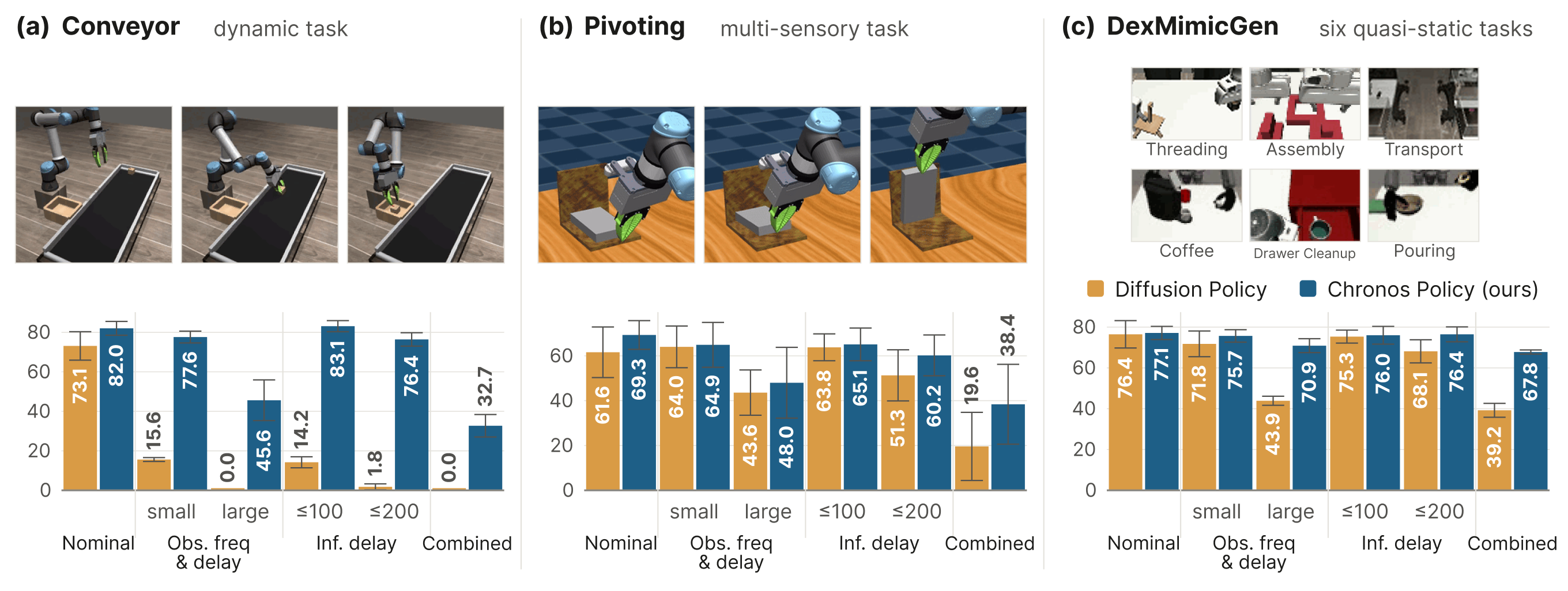}
\caption{\textbf{Simulation evaluation across manipulation settings.}
We compare Diffusion Policy and \methodname{} on (a) dynamic Conveyor, (b) multisensory Pivoting, and (c) six standard DexMimicGen tasks under nominal and perturbed sensing and inference timing.}
\label{fig:simulation}
\end{figure*}

\paragraph{Time substantially improves dynamic manipulation.}
Figure~\ref{fig:simulation}a shows a clear advantage on Conveyor. Under nominal timing, \methodname{} improves success from 73.1\% to 82.0\%. More importantly, the gap grows when deployment timing changes. With moderate observation variation, Diffusion Policy drops to 15.6\% while \methodname{} retains 77.6\%; under inference delays up to 200\,ms, Diffusion Policy falls to 1.8\% while \methodname{} achieves 76.4\%. Even under the combined perturbation, \methodname{} retains 32.7\% success while Diffusion Policy fails entirely.

This result shows that sequence order alone is not enough to determine object motion when observation intervals vary, nor to determine how far ahead to act when execution is delayed. Providing both observation and action times allows the policy to interpret these changes directly.

\paragraph{Time improves robustness in multisensory manipulation.}
The Pivoting results show a different pattern (Fig.~\ref{fig:simulation}b). Under mild timing changes, \methodname{} and Diffusion Policy perform similarly, while the advantage of explicit time becomes larger as temporal mismatch increases. Under inference delays up to 200\,ms, success improves from 51.3\% to 60.2\%, and under the combined condition from 19.6\% to 38.4\%.

Unlike Conveyor, Pivoting does not require continuously predicting the future position of a moving target. Instead, timestamps become useful when the relative times of vision, proprioception, and force/torque measurements vary enough that their sequence positions no longer reliably indicate when they were acquired.

\subsection{What Makes a Policy Time-Aware?}
\label{sec:ablate}

The previous results show that explicit time improves robustness to temporal mismatch. We then study what makes a time-aware policy effective by ablating two key design choices in \methodname{}: temporal variation during training and the representation of physical time in the transformer (Fig.~\ref{fig:ablation}).

\paragraph{Temporal randomization requires explicit time information.}
Temporal randomization has two effects: it exposes the policy to a wider range of sensing and execution schedules that may occur during deployment, and it allows training to use measurements that would otherwise be discarded when demonstrations are downsampled to a fixed temporal grid. This additional diversity can act as useful data augmentation. On Conveyor, temporal randomization alone increases Diffusion Policy's average success across the five perturbed conditions from 6.3\% to 26.9\%. Thus, even without timestamps, exposing the policy to more temporal variation can improve robustness when the relevant timing structure is relatively simple.

The same strategy does not generalize to multisensory Pivoting. There, temporal randomization reduces the perturbed-condition average from 48.4\% to 21.6\%. Without timestamps, the policy cannot tell what timing produced each sampled history: the same sequence positions may correspond to different physical intervals, and asynchronous measurements from different modalities become difficult to interpret. Once physical time is explicit, however, temporal randomization becomes consistently useful. It increases the perturbed-condition average from 48.8\% to 63.1\% on Conveyor and from 33.3\% to 55.3\% on Pivoting. Temporal randomization therefore broadens the training distribution, while timestamps make that variation interpretable.

\paragraph{Time-aware RoPE provides a better representation of physical time.}
We compare \texttt{Chronos-RoPE} with \texttt{Chronos-AddTime}, which follows the common choice of encoding continuous time with Fourier features~\citep{tancik2020fourier}. Each timestamp, expressed relative to the policy request time, is mapped through Fourier features and an MLP, and the resulting time embedding is added directly to the corresponding token. Observation tokens use their capture times, while action tokens use their queried execution times; for action tokens, the time embedding is added to the noisy action representation at each diffusion step. In contrast, \texttt{Chronos-RoPE} keeps the token embeddings unchanged and introduces the same physical timestamps only through the positional encoding in attention.

On Conveyor, \texttt{Chronos-AddTime} performs slightly better under nominal timing, but \texttt{Chronos-RoPE} performs better in every other setting, increasing the temporal condition average from 51.5\% to 63.1\%. On Pivoting, \texttt{Chronos-RoPE} outperforms \texttt{Chronos-AddTime} in every condition and improves the perturbed-condition average from 33.5\% to 55.3\%. We hypothesize this is due to its simpler separation of content and time, and to the fact that pairwise elapsed time directly modulates attention rather than being inferred from additive token features. We therefore use \texttt{Chronos-RoPE} as \methodname{} in the experiments.

\subsection{Does Time Hurt Standard Quasi-Static Manipulation?}

We next evaluate six standard quasi-static manipulation tasks, where precise timing is less critical to the task. Under nominal timing, \methodname{} and Diffusion Policy achieve comparable performance as shown in Fig.~\ref{fig:simulation}c (full results are shown in the Appendix), showing that explicitly modeling time does not sacrifice performance.

The advantage becomes clear when deployment timing differs from training. Averaged across the five perturbed conditions, success improves from 59.7\% with Diffusion Policy to 73.4\% with \methodname{}, with the largest gains under wide sensing variation and combined perturbations.

Overall, \methodname{} preserves nominal performance while providing substantially greater robustness to temporal mismatch. When timing is close to the training setup, explicit time does not sacrifice any performance; when sensing or inference timing changes, it gives the policy information about the observation and action timing that is absent from the baseline.

\subsection{Does Time Awareness Improve Real-World Robustness?}
\label{sec:real}

\begin{figure*}[t]
\centering
\includegraphics[width=\textwidth]{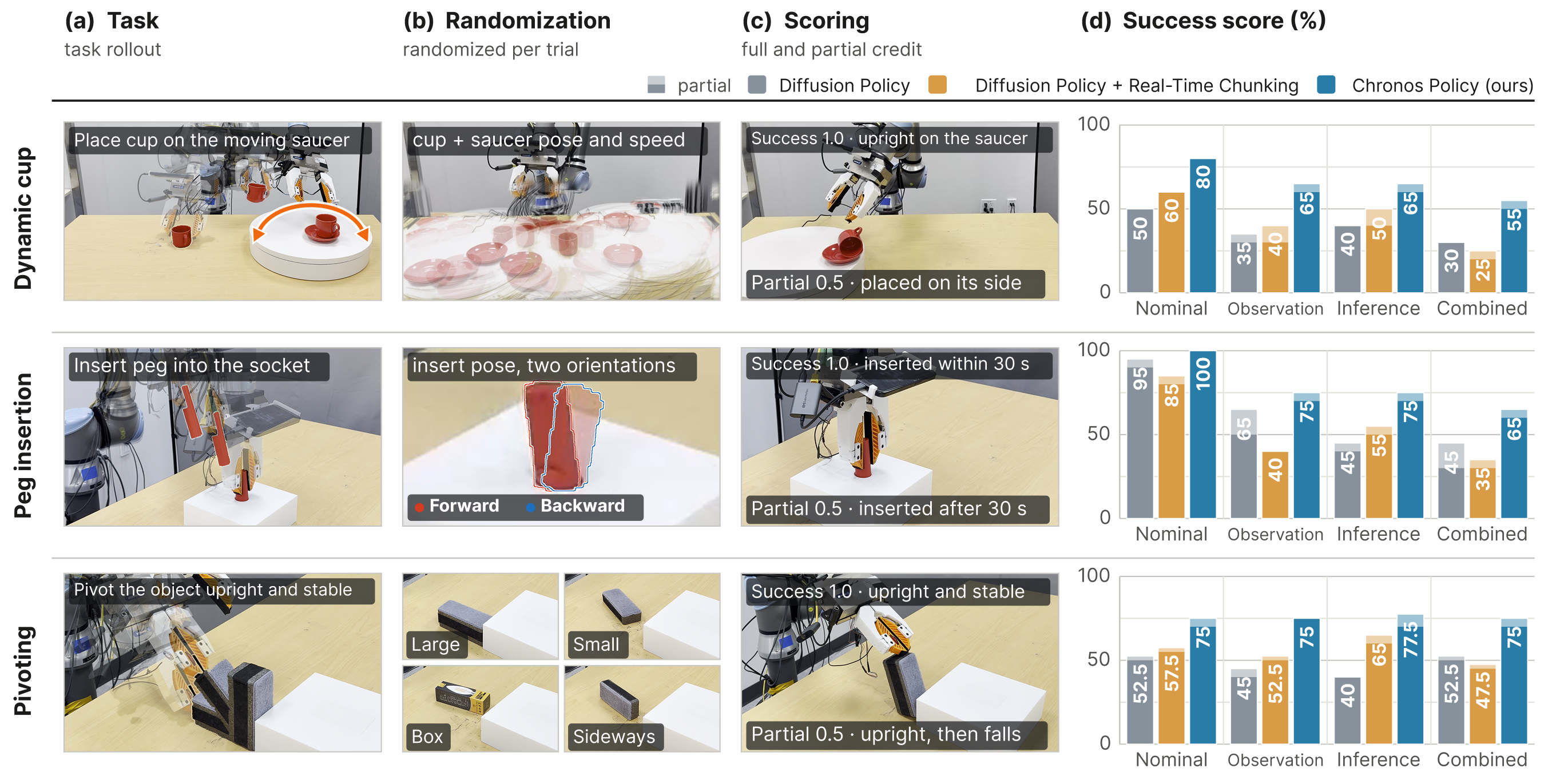}
\caption{\textbf{Real-world evaluation under timing variation.}
We compare Diffusion Policy, Diffusion Policy with RTC, and \methodname{} on Dynamic Cup, Peg Insertion, and Pivoting under nominal, observation, inference, and combined timing conditions. Lighter bar segments indicate partial success.}
\label{fig:realworld}
\end{figure*}

We finally ask whether the benefits of explicit time carry over to real robot deployment, where sensing and inference timing vary naturally. As shown in Figure~\ref{fig:realworld}, we evaluate three tasks that mirror the two settings studied in simulation: \textbf{Dynamic Cup} is a dynamic manipulation task, similar to Conveyor, where the robot must place a cup onto a moving saucer; \textbf{Peg Insertion} and \textbf{Pivoting} are contact-rich multisensory tasks that combine vision, proprioception, and force/torque feedback. We compare asynchronous Diffusion Policy, Diffusion Policy with Real-Time Chunking (RTC), and \methodname{}. Under asynchronous updates, the robot continues executing the previous action chunk while the next prediction is computed. Both RTC and \methodname{} estimate inference latency from previously completed requests.

\methodname{} achieves the highest task score in every evaluated condition across all three tasks. The advantage persists under observation perturbations, inference latency, and their combination, showing that the benefit of explicit time extends beyond controlled simulation. The gains also span both moving-target and contact-rich manipulation, suggesting that the same physical-time interface can handle different sources of temporal variation.

RTC improves robustness to inference latency in several settings. However, RTC does not explicitly account for changes in observation timing, so its benefit does not consistently extend to sensing perturbations. In contrast, \methodname{} represents both when each observation was captured and when each predicted action should execute, allowing sensing and execution timing to be modeled within the same policy architecture. These results suggest that asynchronous execution and explicit time awareness address complementary aspects of temporal mismatch.

\section{Conclusion and Future Work}

We introduce \methodname{}, which grounds robot policy learning in physical time by conditioning on when observations are captured and when actions should execute. This formulation provides a unified representation of sensing rate, measurement delay, and inference latency within a single policy.  Across dynamic, multisensory, and standard quasi-static manipulation tasks in simulation and on real robots, \methodname{} improves robustness to temporal mismatch while preserving nominal performance.

A direction for future work is to extend Chronos policies beyond fixed action rates to variable control frequencies and temporal scaling, allowing robots to execute the same behavior faster or slower than demonstration.

\section{Acknowledgment}
We thank all members of REALab at Stanford for helpful discussions and feedback. This work was supported in part by the Knight-Hennessy Fellowship and the NSF Graduate Research Fellowships Program (GRFP),
NSF Award \#2143601, \#2037101, and \#2132519, 
Samsung, Stanford HAI Center. The views and conclusions contained herein are those of the authors and should not be interpreted as necessarily representing the official policies, either expressed or implied, of the sponsors.

\clearpage
\bibliographystyle{iclr2027_conference}
\bibliography{reference}

\clearpage
\appendix
\raggedbottom

\makeatletter
\setlength{\@fptop}{0pt}
\setlength{\@fpsep}{18pt}
\makeatother

\section{Implementation Details}
\label{app:implementation}

\subsection{Policy architecture and input representations}

\begin{figure}[!htbp]
\centering
\includegraphics[width=\textwidth]{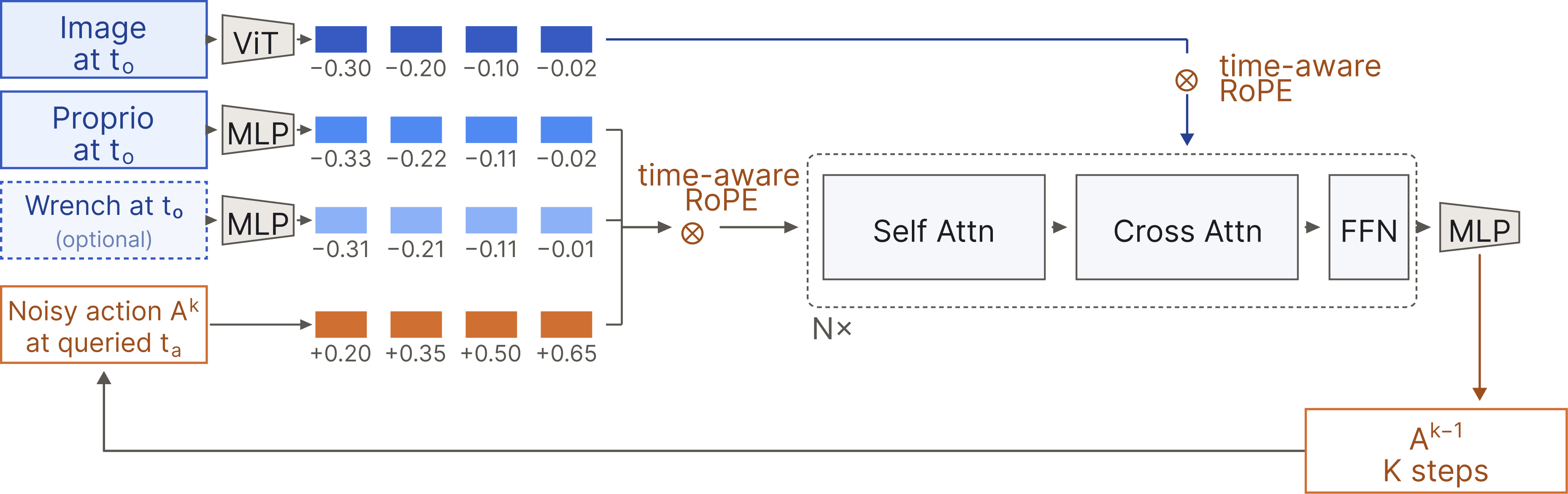}
\caption{\textbf{Architecture of \methodname{}.} Proprioceptive, optional force/torque, and noisy action tokens interact through self-attention and attend to image features through cross-attention. Time-aware RoPE encodes observation capture times and queried action times in attention. The transformer iteratively denoises the action chunk.}
\label{fig:app_architecture}
\end{figure}

Our implementation of Diffusion Policy (DP) and \methodname{} uses the same backbone: a 16-layer diffusion transformer with width 512 and eight attention heads. A pretrained SigLIP2 base encoder extracts 768-dimensional image-patch features at $256\times256$ resolution, which are projected into the transformer. Proprioceptive and action vectors are embedded as separate tokens; Pivoting and Insertion tasks additionally include force/torque tokens. Transformer blocks combine self-attention over non-image tokens with cross-attention to image patches (Fig.~\ref{fig:app_architecture}).

Conveyor and the six standard tasks use two image observations per camera and two proprioceptive observations. Simulated Pivoting uses two images, three pose observations, and 32 six-dimensional force/torque samples. We represent a pose by its position and 6D rotation. Pivoting predicts a 19-dimensional adaptive-compliance action containing a nominal pose, a virtual-target pose, and stiffness following~\citep{hou2025adaptive}; the same controller and action representation are used for all methods. Other simulation tasks use absolute pose and gripper or hand commands. Every policy predicts 16 actions at 20 Hz.

\subsection{Baselines and time-aware variants}
\label{app:time_encoding}

\paragraph{Diffusion Policy.}
The transformer-based baseline uses sequence position without observation or action timestamps. Its temporal-randomization variant uses the same randomized observation schedules and action-start delays as \methodname{}, but does not receive their physical times.

\paragraph{\methodname{}.}
Time-aware RoPE uses physical timestamps in seconds, measured relative to the policy request time. Observation tokens use their capture times, while action tokens use their queried execution times, so inference latency shifts the action timestamps into the future. We rotate 16 query--key coordinate pairs using logarithmically spaced periods between 0.1 and 2\,s. Images, proprioception, force/torque, and actions retain their respective capture or query times.

\paragraph{Chronos AddTime.}
The AddTime variant replaces time-aware RoPE with 16 sine and 16 cosine features of the same timestamps. A shared $32\rightarrow512\rightarrow512$ MLP with GELU maps these features into an embedding added to each timed token. For action tokens, this embedding is added at each denoising step. AddTime and RoPE share the same data, token layout, diffusion conditioning, and temporal randomization.

\paragraph{Real-Time Chunking.}
For the real-world baseline, we adapt RTC~\citep{black2026real} to the Diffusion Policy DDIM sampler using action-prefix inpainting. Previously committed actions constrain the next chunk while inference overlaps execution. RTC and \methodname{} estimate inference latency from completed requests; DP and RTC do not receive observation capture timestamps as policy inputs.

\subsection{Training Details}

We use $\ell_1$ loss for action prediction and a cosine diffusion-noise schedule. We optimize with AdamW, learning rate $3\times10^{-4}$, betas $(0.95,0.999)$, and weight decay $10^{-6}$; the pretrained image encoder uses one tenth of the learning rate. Training uses BF16 and a cosine learning-rate schedule. All policies are trained for 50 epochs, where one epoch is one complete pass through the training dataset. Table~\ref{tab:app_implementation} lists the task-specific settings. Non-rotational action and proprioceptive dimensions use range normalization; 6D rotation coordinates are left unchanged. Image augmentation applies random cropping and color jitter, each with probability 0.3, followed by antialiased resizing to the encoder resolution.

\begin{table}[!htbp]
\centering\small
\caption{\textbf{Simulation training and execution settings.} The action prediction horizon is 16 and action frequency is 20\,Hz for every task.}
\label{tab:app_implementation}
\begin{tabular}{@{}lccc@{}}
\toprule
Setting & Standard tasks & Conveyor & Pivoting \\
\midrule
Training epochs & 50 & 50 & 50 \\
Crop resolution & $224\times224$ & $196\times196$ & $196\times196$ \\
DDIM steps: training / inference & 24 / 8 & 24 / 8 & 24 / 8 \\
Action excution horizon & 12 & 8 & 8 \\
\bottomrule
\end{tabular}
\end{table}

For real-world policies, we use clean action prediction with an $\ell_1$ loss and DDIM with 50 training and 16 inference steps. Images use a 212-pixel random crop during training and a center crop at evaluation, resized to $256\times256$. The Pivoting training configuration uses effective batch size 128, AdamW at $3\times10^{-4}$, an encoder learning-rate multiplier of 0.1, and cosine decay with 2,000 warmup updates.

\FloatBarrier
\section{Additional Simulation and Ablation Results}
\label{app:results}

\subsection{Checkpoint Selection and Evaluation Setup}
\label{app:task_selection}
\label{app:protocols}

We train three seeds for each policy. We evaluate checkpoints on the nominal validation setting every five epochs and select the checkpoint with the highest nominal success rate. For the DexMimicGen benchmark, we evaluate 50 episodes per training seed, following the original paper evaluation setup. For the Conveyor and Pivoting ablations in Fig.~\ref{fig:ablation}, each selected training checkpoint is evaluated with three evaluation seeds, with 50 episodes per seed and condition. We first average across the three evaluation seeds for each training seed, and then report the mean and sample SD across the three training seeds, for a total of 450 episodes per method and condition.

\subsection{Temporal settings}

\paragraph{Training.}
All the tasks sample observation rates of 12.5--33.33\,Hz and delivery delays of 0--50\,ms independently across sensor streams, using a 100\,Hz original dataset. For Pivoting, force/torque is sampled from 1--100\,Hz with 0--100\,ms delivery delay. Half of the action chunks start with zero delay; the rest sample a start delay up to 100\,ms.

\paragraph{Evaluation.}
Table~\ref{tab:app_sensor_ranges} specifies observation frequency and delay. The two inference conditions sample an inference delay from 0--100\,ms or 0--200\,ms. Combined uses the large-observation setting and the large inference distribution.

\begin{table}[!htbp]
\centering
\small
\caption{\textbf{Simulation observation-timing conditions.}
Sensor frequencies are in Hz and delivery delays are in ms.
Small and Large correspond to the observation perturbations used in the main figures; sensor streams are sampled independently.}
\label{tab:app_sensor_ranges}
\setlength{\tabcolsep}{6pt}
\renewcommand{\arraystretch}{1.1}
\begin{tabular}{@{}lcccc@{}}
\toprule
Condition & Image & Pose & Force/torque & Delay \\
\midrule
Nominal & 20 & 20 & 100 & 0 \\
Small   & 12.5--33.33 & 12.5--33.33 & 75--100 & 0--50 \\
Large   & 1--50 & 1--50 & 1--100 & 0--100 \\
\bottomrule
\end{tabular}
\end{table}

\subsection{Dynamic and multisensory tasks: full ablations}
\label{app:ablation}

Table~\ref{tab:app_ablation} provides the full ablation results behind Fig.~\ref{fig:ablation}.

\paragraph{Temporal randomization alone can improve robustness, but does not resolve temporal ambiguity.}
On Conveyor, adding temporal randomization to Diffusion Policy substantially improves several perturbation settings. Success increases from 15.6\% to 60.9\% under Small observation variation and from 14.2\% to 47.3\% under delays up to 100\,ms. This suggests that exposing the policy to a broader range of observation and execution schedules can itself act as useful data augmentation. However, the benefit does not extend cleanly to more severe shifts: under Wide sensing, success only improves from 0.0\% to 6.4\%, and both variants remain near zero under the Combined condition. Randomization also lowers nominal Conveyor performance from 73.1\% to 53.6\%, indicating a trade-off when temporal variation is introduced without telling the policy what timing generated each sample.

This limitation is more pronounced on Pivoting. Adding temporal randomization to Diffusion Policy reduces performance in every condition, including nominal success from 61.6\% to 29.8\%. Unlike Conveyor, Pivoting requires relating vision, proprioception, and force/torque measurements sampled at different times. When these schedules are randomized but their timestamps are hidden, the same token positions can correspond to different cross-modal time offsets. The additional diversity therefore also introduces ambiguity that the policy cannot resolve.

\paragraph{Timestamps and temporal randomization are complementary.}
Explicit time without temporal randomization already provides substantial robustness to execution delay. On Conveyor, RoPE without randomization reaches 76.4\% and 69.3\% under delays up to 100 and 200\,ms, respectively. However, it is much less robust when observation timing changes, dropping to 20.2\% under Wide sensing and 10.7\% under the Combined condition. Adding temporal randomization raises these results to 45.6\% and 32.7\%, respectively. A similar pattern appears on Pivoting: RoPE without randomization drops to 9.1\% under Wide sensing and 4.2\% under Combined, whereas the full model reaches 48.0\% and 38.4\%.

These results clarify the roles of the two components. Timestamps make a particular sensing and execution schedule interpretable, while temporal randomization exposes the policy to a broader set of such schedules during training. Neither component alone provides the same robustness as their combination.

\paragraph{The representation of time also matters.}
AddTime and RoPE receive the same timestamps and temporal randomization, isolating how physical time enters the transformer. AddTime achieves slightly higher nominal performance on Conveyor (85.8\% versus 82.0\%), but RoPE performs better in every perturbed condition. The difference is especially large under Wide sensing (45.6\% versus 24.0\%) and Combined perturbations (32.7\% versus 7.8\%). On Pivoting, RoPE has higher mean success than AddTime in every condition, including 48.0\% versus 23.6\% under Wide sensing and 38.4\% versus 16.7\% under Combined. Since the two variants differ only in how the same physical timestamps are represented, these results are consistent with directly encoding pairwise elapsed time in attention being more robust than adding time to token content.

\begin{table}[!htbp]
\centering\small
\caption{\textbf{Full Conveyor and Pivoting results and design ablations.}
Success rate (\%), mean $\pm$ sample SD across three training seeds after averaging three paired evaluation batches within each seed. Each condition contains 450 episodes per method. RoPE is the full \methodname{}; AddTime changes only the time encoding. Rand. denotes temporal randomization. Best mean performance within each task and condition is shown in \textbf{bold}.}
\label{tab:app_ablation}
\begingroup
\setlength{\tabcolsep}{2.8pt}
\renewcommand{\arraystretch}{1.12}
\begin{tabular}{@{}lcccccc@{}}
\toprule
Method &
Nominal &
Small obs. &
\shortstack{Wide\\sensors} &
\shortstack{Delay\\0--100\,ms} &
\shortstack{Delay\\0--200\,ms} &
Combined \\
\midrule
\multicolumn{7}{@{}l}{\textbf{Conveyor}} \\
DP
& $73.1 \pm 7.2$
& $15.6 \pm 1.0$
& $0.0 \pm 0.0$
& $14.2 \pm 2.8$
& $1.8 \pm 1.5$
& $0.0 \pm 0.0$ \\
DP + rand.
& $53.6 \pm 2.3$
& $60.9 \pm 3.9$
& $6.4 \pm 2.1$
& $47.3 \pm 2.4$
& $19.6 \pm 2.7$
& $0.2 \pm 0.4$ \\
RoPE w/o rand.
& $79.8 \pm 4.2$
& $67.1 \pm 3.4$
& $20.2 \pm 0.8$
& $76.4 \pm 8.6$
& $69.3 \pm 3.1$
& $10.7 \pm 0.7$ \\
AddTime
& $\mathbf{85.8 \pm 7.6}$
& $74.9 \pm 4.5$
& $24.0 \pm 1.3$
& $78.9 \pm 4.4$
& $71.8 \pm 3.4$
& $7.8 \pm 2.0$ \\
RoPE (ours)
& $82.0 \pm 3.5$
& $\mathbf{77.6 \pm 3.0}$
& $\mathbf{45.6 \pm 10.2}$
& $\mathbf{83.1 \pm 2.8}$
& $\mathbf{76.4 \pm 3.4}$
& $\mathbf{32.7 \pm 5.7}$ \\
\midrule
\multicolumn{7}{@{}l}{\textbf{Pivoting}} \\
DP
& $61.6 \pm 11.3$
& $64.0 \pm 9.3$
& $43.6 \pm 10.1$
& $63.8 \pm 6.0$
& $51.3 \pm 11.4$
& $19.6 \pm 15.2$ \\
DP + rand.
& $29.8 \pm 14.5$
& $30.0 \pm 10.6$
& $18.0 \pm 9.3$
& $29.1 \pm 9.1$
& $22.0 \pm 14.0$
& $8.9 \pm 5.4$ \\
RoPE w/o rand.
& $60.9 \pm 5.0$
& $54.0 \pm 10.0$
& $9.1 \pm 8.1$
& $53.1 \pm 7.5$
& $46.2 \pm 11.0$
& $4.2 \pm 4.3$ \\
AddTime
& $50.4 \pm 19.9$
& $46.2 \pm 24.2$
& $23.6 \pm 15.7$
& $40.9 \pm 23.8$
& $40.0 \pm 25.1$
& $16.7 \pm 12.2$ \\
RoPE (ours)
& $\mathbf{69.3 \pm 6.4}$
& $\mathbf{64.9 \pm 10.0}$
& $\mathbf{48.0 \pm 15.8}$
& $\mathbf{65.1 \pm 7.3}$
& $\mathbf{60.2 \pm 9.1}$
& $\mathbf{38.4 \pm 17.8}$ \\
\bottomrule
\end{tabular}
\endgroup
\end{table}

\subsection{Standard manipulation: task-level results}

Table~\ref{tab:app_simulation} reports the task-level results in Fig.~\ref{fig:simulation}c. On average, \methodname{} achieves comparable performance under nominal timing and better performance under temporal perturbations, with the clearest gains on Assembly, Coffee, and Pouring, especially under observation and combined timing changes.

\begin{table}[!htbp]
\centering\small
\caption{\textbf{Full results on the six standard manipulation tasks.} Success rate (\%), mean $\pm$ sample SD over three training seeds. Each seed is evaluated for 50 episodes per condition. DP denotes the transformer-based Diffusion Policy baseline. Best mean success within each task and condition is shown in \textbf{bold}.}
\label{tab:app_simulation}
\begingroup
\setlength{\tabcolsep}{3pt}
\renewcommand{\arraystretch}{1.12}
\begin{tabular}{@{}lcccccc@{}}
\toprule
Method & Nominal & Small obs. & \shortstack{Wide\\sensors} & \shortstack{Delay\\0--100\,ms} & \shortstack{Delay\\0--200\,ms} & Combined \\
\midrule
\multicolumn{7}{@{}l}{\textbf{Threading}} \\
DP & $\mathbf{89.3 \pm 8.3}$ & $88.7 \pm 5.0$ & $\mathbf{78.7 \pm 11.0}$ & $\mathbf{90.0 \pm 7.2}$ & $79.3 \pm 5.0$ & $\mathbf{77.3 \pm 8.1}$ \\
\methodshort{} & $88.7 \pm 3.1$ & $\mathbf{90.0 \pm 2.0}$ & $77.3 \pm 3.1$ & $84.7 \pm 5.0$ & $\mathbf{86.7 \pm 3.1}$ & $63.3 \pm 3.1$ \\
\midrule
\multicolumn{7}{@{}l}{\textbf{Assembly}} \\
DP & $\mathbf{75.3 \pm 7.0}$ & $\mathbf{73.3 \pm 5.0}$ & $13.3 \pm 6.4$ & $\mathbf{72.7 \pm 7.6}$ & $\mathbf{72.7 \pm 1.2}$ & $0.0 \pm 0.0$ \\
\methodshort{} & $68.7 \pm 1.2$ & $\mathbf{73.3 \pm 2.3}$ & $\mathbf{61.3 \pm 3.1}$ & $71.3 \pm 6.1$ & $71.3 \pm 6.4$ & $\mathbf{57.3 \pm 7.6}$ \\
\midrule
\multicolumn{7}{@{}l}{\textbf{Transport}} \\
DP & $84.0 \pm 5.3$ & $\mathbf{81.3 \pm 5.0}$ & $\mathbf{78.0 \pm 5.3}$ & $\mathbf{87.3 \pm 9.9}$ & $\mathbf{84.7 \pm 7.6}$ & $\mathbf{73.3 \pm 8.1}$ \\
\methodshort{} & $\mathbf{87.3 \pm 7.0}$ & $\mathbf{81.3 \pm 11.0}$ & $77.3 \pm 6.1$ & $85.3 \pm 9.0$ & $84.0 \pm 6.0$ & $70.0 \pm 10.4$ \\
\midrule
\multicolumn{7}{@{}l}{\textbf{Coffee}} \\
DP & $47.3 \pm 6.1$ & $34.7 \pm 6.1$ & $14.7 \pm 1.2$ & $41.3 \pm 14.2$ & $27.3 \pm 7.6$ & $2.7 \pm 3.1$ \\
\methodshort{} & $\mathbf{51.3 \pm 14.0}$ & $\mathbf{58.0 \pm 12.2}$ & $\mathbf{57.3 \pm 9.0}$ & $\mathbf{52.7 \pm 15.1}$ & $\mathbf{57.3 \pm 14.5}$ & $\mathbf{60.0 \pm 5.3}$ \\
\midrule
\multicolumn{7}{@{}l}{\textbf{Drawer Cleanup}} \\
DP & $83.3 \pm 6.1$ & $\mathbf{81.3 \pm 4.2}$ & $76.7 \pm 7.0$ & $82.0 \pm 4.0$ & $80.7 \pm 1.2$ & $\mathbf{79.3 \pm 9.0}$ \\
\methodshort{} & $\mathbf{87.3 \pm 6.4}$ & $79.3 \pm 13.3$ & $\mathbf{82.0 \pm 14.4}$ & $\mathbf{85.3 \pm 8.1}$ & $\mathbf{84.7 \pm 8.1}$ & $78.7 \pm 8.1$ \\
\midrule
\multicolumn{7}{@{}l}{\textbf{Pouring}} \\
DP & $\mathbf{79.3 \pm 29.1}$ & $71.3 \pm 38.2$ & $2.0 \pm 2.0$ & $\mathbf{78.7 \pm 30.3}$ & $64.0 \pm 36.7$ & $2.7 \pm 1.2$ \\
\methodshort{} & $\mathbf{79.3 \pm 6.4}$ & $\mathbf{72.0 \pm 6.9}$ & $\mathbf{70.0 \pm 3.5}$ & $76.7 \pm 4.2$ & $\mathbf{74.7 \pm 1.2}$ & $\mathbf{77.3 \pm 5.0}$ \\
\bottomrule
\end{tabular}
\endgroup
\end{table}

\FloatBarrier
\section{Additional Real-World Setup and Results}
\label{app}
\label{app:realworld}

\subsection{Task setup and demonstrations}

We evaluate three real-world tasks: Dynamic Cup, Peg Insertion, and Pivoting. Table~\ref{tab:app_realworld_setup} summarizes the number of demonstrations and the success criteria for each task. Dynamic Cup is evaluated across different saucer speeds, positions, and rotation directions. Peg Insertion is evaluated with five trials for each of two hole orientations, forward and backward. Pivoting is evaluated with five trials in each of four settings: a large eraser, a small eraser, a box, and a sideways configuration.

\begin{table}[!htbp]
\centering\small
\caption{\textbf{Real-world datasets and scoring.} Full success receives one credit, partial success half a credit, and failure zero. Demonstration counts are training / validation.}
\label{tab:app_realworld_setup}
\setlength{\tabcolsep}{4pt}
\begin{tabular}{@{}lp{1.5cm}p{3.5cm}p{3.5cm}@{}}
\toprule
Task & Demos & Full success & Partial success \\
\midrule
Dynamic Cup & 252 / 13 & Cup upright on the moving saucer & Cup placed on its side on the saucer \\
Peg Insertion & 220 / 11 & Peg inserted within 30\,s & Peg inserted after 30\,s \\
Pivoting & 322 / 17 & Object brought upright and stable & Object reaches upright, then falls \\
\bottomrule
\end{tabular}
\end{table}

All tasks use two RGB observations together with histories of relative pose and measured gripper width at a nominal rate of 20\,Hz. Peg Insertion and Pivoting additionally use 32 six-dimensional force/torque measurements from each finger. The force/torque streams are recorded at approximately 360\,Hz and sampled with stride four, giving a nominal history rate of approximately 90\,Hz.

\subsection{Temporal settings and execution}

Evaluation observation settings are listed in Table~\ref{tab:app_realworld_timing}. The inference condition adds a delay sampled from $\{0,50,100,150,200\}$\,ms. The Combined condition applies both the observation perturbation and the inference delay. Nominal uses no injected perturbations, while retaining the system's natural sensing and inference latency.

\begin{table}[!htbp]
\centering\small
\caption{\textbf{Real-world observation-timing settings.} Each stream has an independent delivery delay. Force/torque applies to Peg Insertion and Pivoting.}
\label{tab:app_realworld_timing}
\begin{tabular}{@{}lccc@{}}
\toprule
Stream & Nominal rate (Hz) & Perturbed rate (Hz) & Added delay (ms) \\
\midrule
RGB & 20 & 12.5--33.33 & 0--100 \\
Pose / gripper & 20 & 8.33--33.33 & 0--100 \\
Force/torque & 90 & 37.5--150 & 0--100 \\
\bottomrule
\end{tabular}
\end{table}

\subsection{Full results and analysis}
\label{app:hardware_counts}

For $N$ trials with $F$ full and $P$ partial successes, the task score is
$100(F+0.5P)/N$. Tables~\ref{tab:hardware_cup}--\ref{tab:hardware_pivoting}
report detailed results, with each cell showing task score (\%) followed
by full/partial/failure counts.

\begin{table*}[!htbp]
\centering\small
\caption{\textbf{Dynamic Cup results by motion direction.}
Each direction contains five trials; Overall aggregates all 10 trials per
policy and timing condition. Each cell reports task score (\%) followed by
full/partial/failure counts.}
\label{tab:hardware_cup}
\setlength{\tabcolsep}{5pt}
\renewcommand{\arraystretch}{1.12}
\begin{tabular}{@{}llcccc@{}}
\toprule
Policy & Scenario & Nominal & Observation & Inference & Combined \\
\midrule
DP
& Clockwise
& 40.0 (2/0/3) & 50.0 (2/1/2) & 20.0 (1/0/4) & 40.0 (2/0/3) \\
& Counterclockwise
& 60.0 (3/0/2) & 20.0 (1/0/4) & 60.0 (3/0/2) & 20.0 (1/0/4) \\
& \textbf{Overall}
& \textbf{50.0 (5/0/5)} & \textbf{35.0 (3/1/6)}
& \textbf{40.0 (4/0/6)} & \textbf{30.0 (3/0/7)} \\
\midrule
DP + RTC
& Clockwise
& 60.0 (3/0/2) & 30.0 (1/1/3) & 60.0 (2/2/1) & 20.0 (1/0/4) \\
& Counterclockwise
& 60.0 (3/0/2) & 50.0 (2/1/2) & 40.0 (2/0/3) & 30.0 (1/1/3) \\
& \textbf{Overall}
& \textbf{60.0 (6/0/4)} & \textbf{40.0 (3/2/5)}
& \textbf{50.0 (4/2/4)} & \textbf{25.0 (2/1/7)} \\
\midrule
\methodname{}
& Clockwise
& 80.0 (4/0/1) & 80.0 (4/0/1) & 70.0 (3/1/1) & 50.0 (2/1/2) \\
& Counterclockwise
& 80.0 (4/0/1) & 50.0 (2/1/2) & 60.0 (3/0/2) & 60.0 (3/0/2) \\
& \textbf{Overall}
& \textbf{80.0 (8/0/2)} & \textbf{65.0 (6/1/3)}
& \textbf{65.0 (6/1/3)} & \textbf{55.0 (5/1/4)} \\
\bottomrule
\end{tabular}
\end{table*}

\begin{table*}[!htbp]
\centering\small
\caption{\textbf{Peg Insertion results by hole orientation.}
Forward and backward orientations each contain five trials; Overall aggregates
all 10 trials. Each cell reports task score (\%) followed by
full/partial/failure counts.}
\label{tab:hardware_peg}
\setlength{\tabcolsep}{5pt}
\renewcommand{\arraystretch}{1.12}
\begin{tabular}{@{}llcccc@{}}
\toprule
Policy & Scenario & Nominal & Observation & Inference & Combined \\
\midrule
DP
& Forward
& 90.0 (4/1/0) & 50.0 (2/1/2) & 50.0 (2/1/2) & 50.0 (2/1/2) \\
& Backward
& 100.0 (5/0/0) & 80.0 (3/2/0) & 40.0 (2/0/3) & 40.0 (1/2/2) \\
& \textbf{Overall}
& \textbf{95.0 (9/1/0)} & \textbf{65.0 (5/3/2)}
& \textbf{45.0 (4/1/5)} & \textbf{45.0 (3/3/4)} \\
\midrule
DP + RTC
& Forward
& 80.0 (4/0/1) & 20.0 (1/0/4) & 50.0 (2/1/2) & 70.0 (3/1/1) \\
& Backward
& 90.0 (4/1/0) & 60.0 (3/0/2) & 60.0 (3/0/2) & 0.0 (0/0/5) \\
& \textbf{Overall}
& \textbf{85.0 (8/1/1)} & \textbf{40.0 (4/0/6)}
& \textbf{55.0 (5/1/4)} & \textbf{35.0 (3/1/6)} \\
\midrule
\methodname{}
& Forward
& 100.0 (5/0/0) & 70.0 (3/1/1) & 80.0 (4/0/1) & 70.0 (3/1/1) \\
& Backward
& 100.0 (5/0/0) & 80.0 (4/0/1) & 70.0 (3/1/1) & 60.0 (3/0/2) \\
& \textbf{Overall}
& \textbf{100.0 (10/0/0)} & \textbf{75.0 (7/1/2)}
& \textbf{75.0 (7/1/2)} & \textbf{65.0 (6/1/3)} \\
\bottomrule
\end{tabular}
\end{table*}

\begin{table*}[!htbp]
\centering\small
\caption{\textbf{Pivoting results by object and generalization setting.}
Large and small erasers are in-distribution (ID); the box and sideways setting
are out-of-distribution (OOD). Each individual scenario contains five trials,
and Overall aggregates all 20 trials. Each cell reports task score (\%) followed by
full/partial/failure counts.}
\label{tab:hardware_pivoting}
\setlength{\tabcolsep}{4pt}
\renewcommand{\arraystretch}{1.10}
\begin{tabular}{@{}llcccc@{}}
\toprule
Policy & Scenario & Nominal & Observation & Inference & Combined \\
\midrule
DP
& Large eraser (ID)
& 100.0 (5/0/0) & 100.0 (5/0/0) & 80.0 (4/0/1) & 100.0 (5/0/0) \\
& Small eraser (ID)
& 60.0 (3/0/2) & 60.0 (3/0/2) & 60.0 (3/0/2) & 60.0 (3/0/2) \\
& Box (OOD)
& 40.0 (2/0/3) & 20.0 (0/2/3) & 20.0 (1/0/4) & 40.0 (2/0/3) \\
& Sideways (OOD)
& 10.0 (0/1/4) & 0.0 (0/0/5) & 0.0 (0/0/5) & 10.0 (0/1/4) \\
& \textbf{Overall}
& \textbf{52.5 (10/1/9)} & \textbf{45.0 (8/2/10)}
& \textbf{40.0 (8/0/12)} & \textbf{52.5 (10/1/9)} \\
\midrule
DP + RTC
& Large eraser (ID)
& 100.0 (5/0/0) & 100.0 (5/0/0) & 100.0 (5/0/0) & 80.0 (4/0/1) \\
& Small eraser (ID)
& 100.0 (5/0/0) & 60.0 (3/0/2) & 80.0 (4/0/1) & 60.0 (3/0/2) \\
& Box (OOD)
& 30.0 (1/1/3) & 40.0 (2/0/3) & 60.0 (3/0/2) & 40.0 (2/0/3) \\
& Sideways (OOD)
& 0.0 (0/0/5) & 10.0 (0/1/4) & 20.0 (0/2/3) & 10.0 (0/1/4) \\
& \textbf{Overall}
& \textbf{57.5 (11/1/8)} & \textbf{52.5 (10/1/9)}
& \textbf{65.0 (12/2/6)} & \textbf{47.5 (9/1/10)} \\
\midrule
\methodname{}
& Large eraser (ID)
& 100.0 (5/0/0) & 100.0 (5/0/0) & 100.0 (5/0/0) & 100.0 (5/0/0) \\
& Small eraser (ID)
& 80.0 (4/0/1) & 100.0 (5/0/0) & 100.0 (5/0/0) & 100.0 (5/0/0) \\
& Box (OOD)
& 60.0 (3/0/2) & 40.0 (2/0/3) & 80.0 (4/0/1) & 60.0 (3/0/2) \\
& Sideways (OOD)
& 60.0 (2/2/1) & 60.0 (3/0/2) & 30.0 (0/3/2) & 40.0 (1/2/2) \\
& \textbf{Overall}
& \textbf{75.0 (14/2/4)} & \textbf{75.0 (15/0/5)}
& \textbf{77.5 (14/3/3)} & \textbf{75.0 (14/2/4)} \\
\bottomrule
\end{tabular}
\end{table*}

\FloatBarrier

\end{document}